# High-parallelism Inception-like Spiking Neural Networks for Unsupervised Feature Learning

Mingyuan Meng[a,b,‡], Xingyu Yang[a,‡], Lei Bi[b], Jinman Kim[b], Shanlin Xiao[a,*], and Zhiyi Yu[a,*]

*[a] School of Electronics and Information Technology, Sun Yat-sen University, Guangzhou, China*
*[b] Biomedical & Multimedia Information Technology Research Group, School of Computer Science, The University of Sydney, Sydney, Australia*

Abstract: Spiking Neural Networks (SNNs) are brain-inspired, event-driven machine learning algorithms that have been widely recognized in producing ultra-high-energy-efficient hardware. Among existing SNNs, unsupervised SNNs based on synaptic plasticity, especially Spike-Timing-Dependent Plasticity (STDP), are considered to have great potential in imitating the learning process of the biological brain. Nevertheless, the existing STDP-based SNNs have limitations in constrained learning capability and/or slow learning speed. Most STDP-based SNNs adopted a slow-learning Fully-Connected (FC) architectures and used a sub-optimal vote-based scheme for spike decoding. In this paper, we overcome these limitations with: 1) a design of high-parallelism network architecture, inspired by the Inception module in Artificial Neural Networks (ANNs); 2) use of a Vote-for-All (VFA) decoding layer as a replacement to the standard vote-based spike decoding scheme, to reduce the information loss in spike decoding and, 3) a proposed adaptive repolarization (resetting) mechanism that accelerates SNNs' learning by enhancing spiking activities. Our experimental results on two established benchmark datasets (MNIST/EMNIST) show that our network architecture resulted in superior performance compared to the widely used FC architecture and a more advanced Locally-Connected (LC) architecture, and that our SNN achieved competitive results with state-of-the-art unsupervised SNNs (95.64%/80.11% accuracy on the MNIST/EMNISE dataset) while having superior learning efficiency and robustness against hardware damage. Our SNN achieved great classification accuracy with only hundreds of training iterations, and random destruction of large numbers of synapses or neurons only led to negligible performance degradation.

## 1 Introduction

Recently, Artificial Neural Networks (ANNs) have made good progress in many cognitive tasks (e.g., recognition, analytics, and inference) [19-21]. However, ANNs are computational-intensive [16], and thus research directions have been redirected to brain-inspired Spiking Neural Networks (SNNs) to reduce the computation cost [2]. Unlike the traditional ANNs whose neurons are characterized by static, continuous-valued activation, SNNs resemble the brain's biological functions by using discrete spikes to compute and transmit information. SNNs are thus arguably the only viable way to understand how the brain computes at the neuronal description level [2]. Besides, the power consumption and latency of SNNs can be significantly reduced, compared to ANNs, due to their event-driven style of computing [16-18].

Most existing SNNs can be divided into three categories: *supervised*, *unsupervised*, and *conversion*. *Supervised/ Unsupervised* denotes the algorithms of training SNNs with/without label information. Specifically, supervised SNNs use a loss function to guide their training, which aims at reducing the differences between output and labels [3-6], while unsupervised SNNs adjust their synaptic weights based on biological synaptic plasticity to learn the inner structures of input unlabeled samples [7-14]. *Conversion* denotes the algorithms of converting a trained ANN into a

This work is supported in part by the grants 2017YFA0206200, 2018YFB2202601 from National Key R&D Program of China, and the grants 61674173, 61834005, and 61902443 from National Natural Science Foundation of China (NSFC).
‡ M. Meng and X. Yang contributed equally to this paper. * S. Xiao and Z. Yu both are corresponding authors.
E-mail: mmen2292@uni.sydney.edu.au (M. Meng), yangxy266@mail2.sysu.edu.cn (X. Yang), lei.bi@sydney.edu.au (L. Bi), jinman.kim@sydney.edu.au (J. Kim), xiaoshlin@mail.sysu.edu.cn (S. Xiao), yuzhiyi@mail.sysu.edu.cn (Z. Yu).

SNN to circumvent the difficulties in training SNNs directly [15-18]. In this paper, we focus on unsupervised SNNs because the unsupervised SNNs based on synaptic plasticity are considered more biologically plausible with a higher resemblance to the learning process of the biological brain [1].

Most existing unsupervised SNNs were trained through competitive learning based on Spike-Timing-Dependent Plasticity (STDP) [7-13]. This STDP-based approach allows SNNs to learn in a fully unsupervised fashion without any label information. Despite this advantage, existing unsupervised SNNs have several limitations: In a preliminary study [7], a Fully-Connected (FC) SNN was proposed which exhibited sub-optimal learning capability (95.00% accuracy on the MNIST) and extremely slow learning speed (900,000 iterations were needed). In later studies [9-10], scholars focused on more advanced learning rules such as Adaptive Synaptic Plasticity (ASP) [9] and stochastic STDP [10]. These works indeed improved SNNs' learning capability, but they are still limited by the slow-learning FC network architecture, and thus resulting in sub-optimal learning speed. More recently, Saunders et al. [8] incorporated a Locally-Connected (LC) network architecture into SNNs (called LC SNNs), which improved SNNs' learning speed (60,000 iterations were needed) and robustness against hardware damage. Nonetheless, the LC SNNs' improvement in learning capability was still marginal (95.07% accuracy on the MNIST). In this paper, we focus on designing a STDP-based SNN that improves in all learning efficiency (speed), robustness, and learning capability.

SNNs' learning efficiency and robustness are highly relevant to their architecture parallelism because a high-parallelism SNN consists of multiple independent sub-networks that can learn and compute in parallel (discussed in Section 7.1). For improving SNNs' architecture parallelism, we were motivated by the Inception module [22] in ANNs. The Inception module uses a Split-and-Merge architecture: The input is split into a few parallel pathways with multi-scale filters (e.g., 3×3, 5×5, 7×7 convolutional kernels, pooling, etc.), and then all pathways are concatenated together. Through this Split-and-Merge architecture, the Inception module can process multi-scale spatial information and improve the network's parallelism. Inspired by this Split-and-Merge architecture, we designed an Inception-like multi-pathway network architecture (Fig. 3). To further improve the architecture's parallelism, we divided each pathway into multiple parallel sub-networks by partitioning each competition area into multiple sub-areas during competitive learning (see Section 4.1).

Most STDP-based SNNs adopted a standard vote-based scheme for spike decoding (see Section 3.4) [7, 9-13]. We, however, found that the standard vote-based spike decoding scheme does not work well with the LC/our network architecture. This is due to an underlying assumption of this standard scheme being violated, and being responsible for a large information loss in spike decoding. The standard vote-based decoding scheme assumes that each output neuron can only respond to one certain class and then builds a Vote-for-One (VFO) relationship between each output neuron and a certain class. However, in the LC/our architecture, the Receptive Fields (RFs) of output neurons are localized, and each output neuron can respond to multiple classes that share the same local features (discussed in Section 7.2). To reduce the information loss in spike decoding, we extended this standard vote-based decoding scheme to a Vote-for-All (VFA) decoding layer by building a VFA relationship between each output neuron and all possible classes (see Section 4.2).

Besides, the learning speed of STDP-based SNNs is highly relevant to their spiking intensity, and this is because the update of synaptic weights (i.e., learning) occurs when neurons fire spikes (discussed in Section 7.3). Based on this relevance, we proposed an adaptive repolarization (resetting) mechanism to enhance SNNs' spiking activities, which thereby allows to accelerate SNN's learning (see Section 4.3).

To sum up, in this paper we proposed a STDP-based unsupervised SNN including three key contributions:

- We designed a high-parallelism Inception-like network architecture for SNNs. This architecture integrates multi-scale spatial information (features) and has high parallelism, thus exhibiting improved learning capability, learning efficiency, and robustness against hardware damage.
- We extended the standard vote-based spike decoding scheme to a VFA decoding layer to reduce the information loss during the spike decoding process.
- We proposed an adaptive repolarization mechanism that can enhance SNNs' spiking intensity to accelerate SNNs' learning.

We demonstrate the improvements to the state-of-the-art by experimenting with well-established hand-written digit/letter classification tasks on two public, well-benchmarked datasets (MNIST [23]/EMNIST [24]).

## 2 Related Works

The Inception module was first proposed by Szegedy et al. [22] and then evolved to many variants [25-26] in ANN literature. The earlier instance of Inception with SNNs was reported by Xing et al. [15]; however, in this study, their SNN needed to be trained through a network conversion (i.e., from trained ANN to SNN). From our review, this is the first study to incorporate the concept of Inception into unsupervised SNNs. This proposed study introduces an Inception-like architecture and demonstrates its superior performance compared to the widely used FC/LC architecture with a preprint in [33]. By using [33] as the baseline, we demonstrated that our Inception-like architecture can be extended to be used as a multi-layer unsupervised SNN [38], and Yang et al. [40] also demonstrated that the proposed adaptive repolarization mechanism can be extended to be a stochastic spiking neuron model.

Diehl et al. [7] proposed one of the earliest studies of training unsupervised SNNs through STDP-based competitive learning, where they used a FC network architecture and a relatively simple learning rule. Their proposed FC architecture was widely used in later studies [9-10, 12-13], but this SNN is limited by sub-optimal learning capability (95.00% accuracy on the MNIST) and extremely low learning efficiency (900,000 training iterations were needed). To overcome these limitations: 1) In the aspect of learning rules, Panda et al. [9] proposed an Adaptive Synaptic Plasticity (ASP), inspired by the 'ability to forget' in the human brain, and She et al. [10] proposed to incorporate stochastic STDP into SNNs. These studies indeed improved SNNs' learning capability, but they are still limited by the slow-learning FC architecture. 2) In the aspect of network architecture, Rathi et al. [11] proposed a STDP-based pruning method to compress the FC architecture into a sparsely connected architecture, and Saunders et al. [8] proposed a LC network architecture as a replacement to the FC architecture and got a LC SNN. These studies improved SNNs' learning efficiency and robustness against hardware damage, but the improvement in learning capability is still marginal. Besides, some scholars focused on the implementations of unsupervised SNN on neuromorphic hardware (e.g., Lammie et al. [12]).

Moreover, motivated by the Convolutional Neural Networks (CNNs) in ANN literature, Convolutional Spiking Neural Networks (CSNNs) were proposed for feature extraction [27-30]. Unlike CNNs that are normally trained through supervised learning, CSNNs allow for unsupervised learning of spiking features. However, CSNNs are usually reliant on external supervised classifiers (e.g., SVM, KNN, etc.) to complete the classification process, and therefore they are often considered as semi-supervised SNNs [8]. Furthermore, brain-inspired reservoir-based SNNs were proposed for feature extraction [41-42]. Compared to widely used layer-based SNNs, these reservoir-based SNNs use a 3D network architecture that is shaped like the primary visual cortex. Nikola K. Kasabov [41] proposed a reservoir-based SNN, named NeuCube, to learn and understand spatio-temporal brain data, in which a recurrent 3D SNN reservoir (SNNr) was used to learn spiking features in an unsupervised fashion. Based on NeuCube, Paulun et al. [42] proposed a SNN for moving objects recognition and achieved a competitive result to layer-based SNNs. Nevertheless, reservoir-based SNNs haven't shown superior learning capability than the layer-based SNNs (e.g., Paulun et al. [42] acknowledged that Stromatias et al. [43] achieved higher accuracy using a semi-supervised layer-based SNN), and they also require external supervised classifiers to complete the classification process.

## 3 Fundamentals

In this section, we describe the fundamental components of SNNs, including spiking neuron models, synapse models, spike coding, and competitive learning implementations. The models/methods described in this section are used in our SNN as default if no extra setting is stated.

### 3.1 Spiking Neuron Model

Spiking neurons are the basic computing units of SNNs. For computational efficiency, Integrate-and-Fire (IF) and Leaky Integrate-and-Fire (LIF) models [39] are widely used due to their simplicity [7-11], but there are still many variations of neuron models which implement more complex biological neuron behaviors. For example, the Hodgkin-Huxley (HH) model proposed by Hodgkin et al. [31] is famous for approximating the electrical characteristics of real biological neurons, but its limitation is that it cannot be used for large scale networks due to its excessive computation load. In another study, the neuron model proposed by Izhikevich [32] can simulate rich firing patterns with less computational load than the HH model; nonetheless, it's still not as widely-used as the IF and LIF models because the simplicity of IF and LIF models is more desirable for SNNs' hardware implementations. In this study, since the LIF model was used in most of the related studies [7-11], we also used a simple LIF model for a fair comparison. Other more advanced spiking neuron models can be considered, but the simple LIF model has worked well (95.64%/80.11% accuracy on the MNIST/EMNISE dataset; see Section 6.1). Following Diehl et al. [7], we give the dynamics of the used LIF model as follows:

$$\tau_v \frac{dv}{dt} = v_{rest} - v + (v_{exc} - v)g_e + (v_{inh} - v)g_i \,, \tag{1}$$

$$\tau_{g_e} \frac{dg_e}{dt} = -g_e + \sum_k^{N_{exc}} w_k F(s_k, t)\, \tau_{g_e} \,, \tag{2}$$

$$\tau_{g_i} \frac{dg_i}{dt} = -g_i + \sum_{k}^{N_{inh}} w_k F(s_k, t) \, \tau_{g_i} \,,\qquad(3)$$

$$F(s,t) = \begin{cases} 1 & when\ a\ spike\ is\ received\ at\ t \\ 0 & else \end{cases}.\qquad(4)$$

In (1), $v$ is the membrane voltage (membrane potential) of this neuron, $v_{rest}$ is the resting voltage, $g_{e/i}$ denotes the total excitatory/inhibitory conductance that is input to the neuron, $v_{exc/inh}$ is the equilibrium voltage of excitatory/inhibitory conductance, and $\tau_{v/g_e/g_i}$ is the time constant of $v/g_e/g_i$. In (2) and (3), $N_{exc/inh}$ is the number of excitatory/inhibitory synapses connected to the neuron, $s_k$ is a synapse, and $w_k$ is the synaptic weight of $s_k$. In (4), the function $F(s,t)$ is equal to 1 when the neuron receives a spike through the synapse $s$ at time $t$, otherwise it's equal to 0. According to (1), (2), (3) and (4), the $v$ and $g_{e/i}$ decay exponentially to $v_{rest}$ and 0, respectively, when no spike arrives. At the occurrence of a spike from an excitatory/inhibitory synapse, the $g_{e/i}$ increases by the weight of this synapse, thus leading to the increase/decrease of the membrane voltage $v$. When the $v$ reaches or exceeds a threshold $v_{thres}$, the neuron fires a spike (i.e., depolarization) to downstream neurons, and the $v$ is reset (i.e., repolarization) to a resetting voltage $v_{reset}$. After firing a spike, the neuron does not integrate input spikes for a refractory period $T_{ref}$.

Besides, the homeostasis mechanism used in [7] is adopted to ensure that no neuron can fire excessive spikes and dominate the firing activities of SNNs. This mechanism is an adaptive threshold scheme as follows:

$$\tau_\theta \frac{d\theta}{dt} = v_{thres} - \theta + \theta_{plus} F_n(t) \, \tau_\theta \,,\qquad(5)$$

$$F_n(t) = \begin{cases} 1 & when\ a\ spike\ is\ fired\ at\ t \\ 0 & else \end{cases},\qquad(6)$$

where $\theta$ is an adaptive threshold, $\tau_\theta$ is the time constant of $\theta$, and function $F_n(t)$ is equal to 1 when the neuron fires a spike at time $t$, otherwise it's equal to 0. According to (5) and (6), each time the neuron fires a spike, the threshold $\theta$ increases by a constant $\theta_{plus}$, or it decays exponentially to the $v_{thres}$.

### 3.2 STDP Synapse Model

A synapse is defined as the connection between two neurons (named pre/postsynaptic neuron). STDP synapse models were widely used in unsupervised SNNs [7-13] to model the behaviors of synaptic weights. They have evolved into many variants such as additive STDP [34], triplet STDP [35], and stochastic STDP [10]. In this study, following [7] and [8], we adopted the triplet STDP and simplify it to:

$$\Delta w = \begin{cases} \eta_{post} x_{pre} x_{post2} & when\ postsynaptic\ spike \\ -\eta_{pre} x_{post1} & when\ presynaptic\ spike \end{cases},\qquad(7)$$

where $x_{pre}$ and $x_{post1/post2}$ are the presynaptic and postsynaptic traces [36], $\eta_{pre}/\eta_{post}$ is the pre/postsynaptic learning rate. The synaptic weight $w$ changes when the pre/postsynaptic neuron fires a spike (named pre/postsynaptic spike). The $x_{pre}$ and $x_{post1/post2}$ are reset to 1, respectively, when presynaptic and postsynaptic spikes are fired, or they decay exponentially to 0 with $\tau_{pre}/\tau_{post1}/\tau_{post2}$ as the time constant of $x_{pre}/x_{post1}/x_{post2}$. The use of these synaptic traces is actually equivalent to recording the time when the last pre/postsynaptic spike is fired. Note that we set $\eta_{post} >> \eta_{pre}$ to emphasize the effects of presynaptic neurons on postsynaptic neurons.

The triplet STDP described in (7) adjusts the synaptic weight based on the relative timing of pre/postsynaptic spikes. Fig.1 is an illustration of the triplet STDP model, in which the $t_{pre}/t_{post}$ denotes the time when a pre/postsynaptic spike is fired. As shown in Fig.1, if presynaptic spikes tend to occur immediately before postsynaptic spikes (Fig.1(a)), the synaptic weight tends to be larger. This is because the $x_{post1}$ has already decayed to a relatively small value when presynaptic spikes trigger the updates of the synaptic weight $w$, while the $x_{pre}$ is still a relatively large value when postsynaptic spikes are fired. Similarly, if presynaptic spikes tend to occur immediately after postsynaptic spikes (Fig.1(b)), the synaptic weight tends to be smaller. Competitive learning was proposed mainly based on the former phenomenon (see Section 3.5), which explains why we set $\eta_{post} >> \eta_{pre}$ to emphasize the former phenomenon. In addition, compared to typical STDP models that only have a single postsynaptic trace, the triplet STDP also considers the time interval of postsynaptic spikes through $x_{post2}$. If the time interval between the current and last postsynaptic spikes is too large, the $x_{post2}$ will decay to a very small value. In this case, it is difficult for the synaptic weight $w$ to increase even when presynaptic spikes occur immediately before postsynaptic spikes.

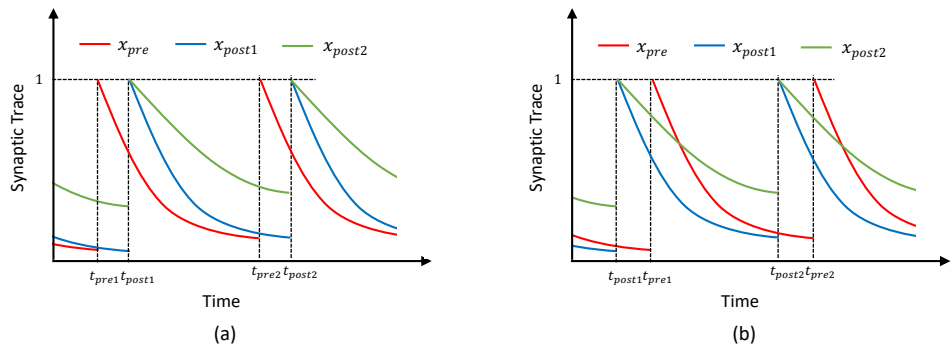

Fig. 1. An illustration of the triplet STDP, which shows the behaviors of synaptic traces when (a) presynaptic spikes tend to occur immediately before postsynaptic spikes, and when (b) presynaptic spikes tend to occur immediately after postsynaptic spikes.

### 3.3 Rate-based Input Encoding

Normally, the natural input of many applications is analog values (e.g., image pixels), but SNNs expect spikes. Therefore, encoding analog values into discrete spike trains is needed. In this study, we adopted a common rate-based encoding scheme that has been widely used in many studies [7-13], where input neurons (i.e., the neurons in the input layer) are generators of Poisson-distributed spike trains. Each input neuron corresponds to a pixel of input images. During encoding, each pixel value is modeled by a Poisson-distributed spike train, and the average rate of the spike train is determined by the pixel value multiplied by an encoding parameter $\lambda$. Besides, since SNNs might be insensitive to some training samples, thus resulting in sub-optimal training with these samples [7], we adopted the adaptive encoding scheme used in [7], in which the $\lambda$ can be adaptive when SNNs' output spiking intensity is too low.

### 3.4 Vote-based Spike Decoding

Since the output of SNNs is spike trains, we need to decode the output spike trains into recognizable results when we apply a trained SNN to a classification task. In other words, we need to construct a mapping from the patterns of output spike trains to the inference results, which is similar to finding the representation of each cluster in K-means clustering. Note that this is the only step where labels are needed.

In general, unsupervised SNNs using rate-based input encoding rely on vote-based methods for spike decoding [7-13]. A standard vote-based spike decoding scheme was widely used in many unsupervised SNNs [7, 9-13]. In this decoding scheme: First, each output neuron (i.e., the neuron in the output layer) is assigned to a certain class based on its highest spiking response, through which a VFO relationship between each output neuron and a certain class is built. Then, during inference, each spike fired by the output neuron is a single vote for its assigned class. Finally, the class having the most votes is the inference result.

The standard vote-based decoding scheme has worked well in most unsupervised SNNs using the FC architecture [7, 9-13], but we found that this decoding scheme is not applicable to the LC architecture and our proposed architecture. This part will be further discussed in Section 7.2.

### 3.5 Competitive Learning

Competitive learning has been widely used as a training approach in many unsupervised SNNs [7-13]. It's based on STDP models and its principle is that: it makes each output neuron compete with each other to learn a certain feature. Each output neuron represents a randomly initialized feature at the beginning. Every time a new training sample comes, only the winner neurons in the competition, whose represented features are the most similar to the features in this training sample, can fire spikes and adjust their synaptic weights based on STDP. In doing so, the winner neurons' represented features will approximate to the real training features. After training, each output neuron will become highly active to different features, and can be used to infer the classes of unseen samples.

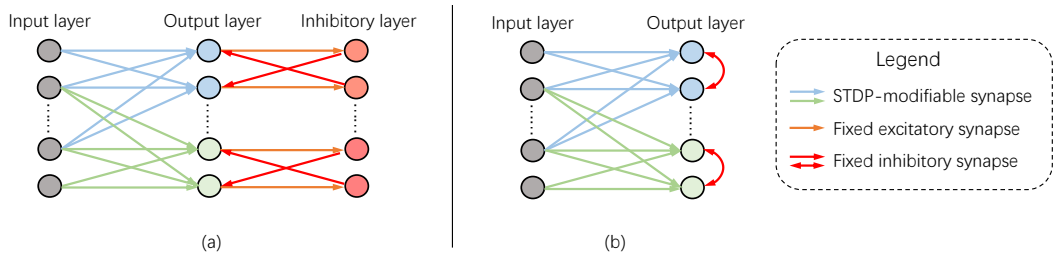

Fig. 2. The implementations of competitive learning. (a) Preliminary version: An inhibitory layer is used to realize competition among output neurons. (b) Alternative version: The inhibitory layer is removed, and the output neurons are directly interconnected with fixed inhibitory synapses. Note that the output neurons in the same competition area are drawn in the same color.

The implementations of competitive learning are shown in Fig. 2: The output layer is connected to the input layer using STDP-modifiable (i.e., modified based on STDP) synapses. Each output neuron competes with other output neurons that are located in the same competition area. Here we define that the output neurons sharing the same RF (i.e., sharing the same set of presynaptic neurons) are in the same competition area. Note that this design is also called lateral inhibition in many studies [7, 9, 11-13]. Fig. 2(a) shows a preliminary version of competitive learning: The output layer is connected to an inhibitory layer using fixed excitatory synapses in a one-to-one manner, and each inhibitory neuron (i.e., the neuron in the inhibitory layer) corresponds to a output neuron. Then, each inhibitory neuron, using fixed inhibitory synapses, is connected to all the output neurons that are located in the same competition area, except for its corresponding one. Besides, there is an alternative version shown in Fig. 2(b): The inhibitory layer is replaced by the inter-connections between output neurons. Each output neuron, using fixed inhibitory synapses, is directly connected to other output neurons that are located in the same competition area. Since the alternative version requires less spiking neurons compared to the preliminary version, it's used in this study as default.

# 4 Method

## 4.1 High-Parallelism Network Architecture

Fig. 3 is an illustration of our proposed network architecture. Inspired by the Inception module in ANN literature, we designed a high-parallelism Inception-like architecture. There are three independent pathways where an output layer is connected to the input layer in a FC or LC manner. The three pathways can compute in parallel without any interaction. A VFA decoding layer is connected to the three output layers in a FC manner to integrate information from the three pathways and then to decode output spike trains into recognizable inference results.

The input neurons are the spike train generators described in Section 3.3, and the output neurons are modeled by the LIF model described in Section 3.1. Following the principle of competitive learning described in Fig. 2(b), the output layers are connected to the input layer using excitatory STDP-modified synapses, and the output neurons in the same competition area (i.e., sharing the same RF) are interconnected using fixed inhibitory synapses.

The LC output layer has the same connection topology as the convolutional layer in ANNs but it does not use shared weights. In this study, only square kernel is used, so we use one kernel size parameter $k$ and one stride parameter $s$ to define RFs (denoted by $(k, s)$ in Fig. 3), and use a parameter $F$ to denote feature map number. Note that the FC output layer actually is a special case of the LC output layer where the kernel size is equal to the size of input layer. Therefore, we also use $k$, $s$, and $F$ to define the FC output layer with $(28,1)$ as $(k, s)$. In this case, each output neuron in the FC layer can be regarded as a feature map. The topology of each pathway is shown in Fig. 3: The RF settings of three pathways are different, which means each pathway can cope with the image features at different scales. The multi-scale features finally merge together in the VFA decoding layer.

Moreover, to further improve the architecture's parallelism, we partition the original competition areas into multiple sub-areas, and only allow the output neurons located in the same sub-area to compete. More specifically, as shown in Fig. 3, the only competition area in the 1st output layer is partitioned into 4 sub-areas, and the 4 competition areas in the 2nd output layer are partitioned into 8 sub-areas, while the 9 competition areas in the 3rd output layer remain unchanged. Finally, we manually set each output layer's $F$ to make sure all sub-areas are equal in size. If we define a parameter $Size_{SA}$ as the size of sub-areas, the $F$ of the 1st/2nd/3rd output layer should be $4Size_{SA}/2Size_{SA}/Size_{SA}$. After this partitioning, our architecture consists of 21 individual competition areas, which is equivalent to having 21 parallel sub-networks. This part will be further discussed in Section 7.1.

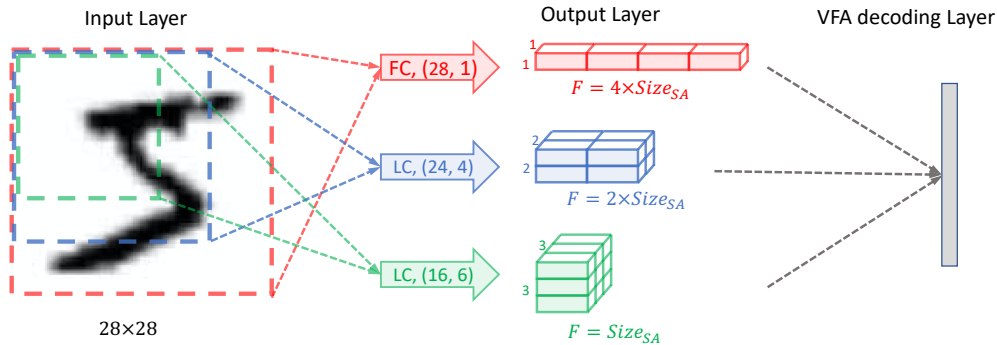

Fig. 3. An illustration of our proposed high-parallelism network architecture. There are three parallel pathways connected to the input layer, each of which has different RF and network topology to cope with image features at different scales. The multi-scale information is finally integrated by a VFA decoding layer. Note that each cube in the output layer denotes a parallel competition area or sub-area.

Note that the design of our architecture is highly empirical. Other designs can be considered if they meet the following principles: 1) There are multiple independent pathways, each of which has different RF setting to cope with the image features at different scales and, 2) The original competition areas can be partitioned into sub-areas, but all sub-areas should be equal in size. This part will be further discussed in Section 7.1.

### 4.2 VFA Decoding Layer

The VFA decoding layer is connected to three pathways in a FC manner to receive the votes from all output neurons. Here the *VFA(Vote-for-all)* means each output neuron needs to vote for all classes. This decoding layer can integrate multi-scale spatial information and then decode output spike trains into recognizable inference results. Each VFA neuron (i.e., the neuron in the VFA decoding layer) corresponds to a class, so the number of VFA neurons is equal to the number of all possible classes in the target classification task. The VFA neurons are modeled by a specially designed voltage-based IF model whose threshold is set to be infinite, meaning that the VFA neurons have no ability to fire spikes and their membrane voltage increases from 0 to infinity when they receive spikes. Actually, each VFA neuron serve as a vote counter to accumulate the votes from the output neurons. During inference, the VFA neuron having the highest membrane voltage indicates the final inference result.

The VFA decoding layer only works during inference and is excluded during training, so its synaptic weights are fixed to be 0 during training. After the training is finished, the synaptic weights of the VFA decoding layer are calculated based on output neuron's spiking response to training samples. Here we define $w_{ij}$ as the synaptic weight between the i[th] output neuron and the j[th] VFA neuron. The $w_{ij}$ can be calculated as follows:

$$w_{ij} = \frac{s_{ij}{}^{\mu}}{\sum_{k=1}^{C} s_{ik}{}^{\mu}},$$  (8)

where $\mu$ is an empirical constant, $C$ is the number of all possible classes, and $s_{ij}$ represents the average number of the spikes that the i[th] output neuron fired to respond to the training samples in the j[th] class.

### 4.3 Adaptive Repolarization

As we mentioned in Section 3.5, the principle of competitive learning is to make the winner neurons in competition adjust their represented features to approximate to real training features. Since the updates of synaptic weights (i.e., learning) are triggered only when spikes are fired, we proposed an adaptive repolarization (resetting) mechanism to enhance the spiking activities of the winner neurons, thereby accelerating competitive learning. This adaptive repolarization mechanism is described as follows:

$$\psi = \begin{cases} v_{reset} + \alpha \Delta C & (\Delta g > 0) \\ v_{reset} - \alpha \Delta C & (\Delta g < 0) \\ v_{reset} & (\Delta g = 0) \end{cases},$$  (9)

$$\Delta g = \left[g_e(t_f) - g_e(t_{f-1} + T_{ref})\right] - \left[g_i(t_f) - g_i(t_{f-1} + T_{ref})\right],$$  (10)

$$\Delta C = v_{thres} - v_{rest},$$  (11)

where $\psi$ is an adaptive resetting voltage, $t_f/t_{f-1}$ denotes the time when this/last spike is fired, $g_{e/i}(t)$ is the value of $g_{e/i}$ at the time $t$, and $\alpha$ is a hyperparameter ranging from 0 to 1. As shown in Fig. 4, the $\psi$ increases by $\alpha \Delta C$ if the neuron gets more excitation than inhibition from $t_{f-1} + T_{ref}$ to $t_f$, and vice versa. Therefore, the spiking activities of winner neurons are enhanced, and other neurons are inhibited.

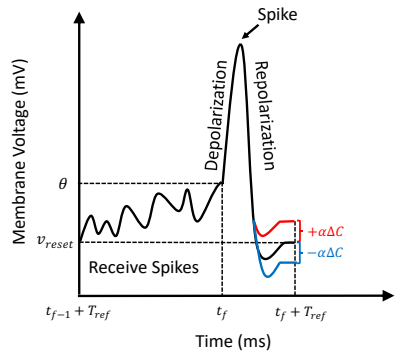

Fig. 4. An illustration of the proposed adaptive repolarization mechanism. The resetting voltage $\psi$ is adaptive based on the spikes received from $t_{f-1} + T_{ref}$ to $t_f$.

Theoretically, a larger $\alpha$ can accelerate SNNs' learning more, but it possibly lets a few neurons fire excessive spikes and dominate the firing activities of the whole competition area. Therefore, there exists a tradeoff between learning efficiency and learning capability. To ease this problem, we set a large $\alpha$ at the beginning of training, and then reduce the $\alpha$ as the training goes, which allows SNNs to converge quickly at the beginning but finally still achieve the same-level learning capability. This part will be further discussed in Section 7.3.

## 5 Experimental Setup

### 5.1 Datasets

- MNIST [23]: The MNIST dataset contains 70,000 hand-written digit images (28×28 in size), split into 60,000 images for training and 10,000 images for testing. These images are labeled into 10 classes from 0 to 9.

- EMNIST [24]: This dataset is an extension of MNIST to hand-written English alphabet letters. There are six partitions in the EMNIST dataset, and we used the letter partition. This partition contains 145,600 hand-written letter images (28×28 in size), split into 124,800 images for training and 20,800 images for testing. These images are labeled into 26 classes corresponding to the 26 capital letters.

### 5.2 Baseline Methods

We evaluated SNNs in terms of learning capability (i.e., classification accuracy), learning efficiency, and robustness against hardware damage. We compared our method with the following baseline methods:

- Diehl-FC [7]: A 3-layer unsupervised FC SNN using a conductance-based LIF model, triplet STDP [35], and the preliminary version of competitive learning (Fig. 2(a)). This SNN has been widely adopted as a baseline method in many studies [8-13]. The public code provided by Diehl et al. [7] was used in the experiments for a fair comparison, which is available at *https://github.com/peter-u-diehl/stdp-mnist*.

- Saunders-LC [8]: A 2-layer unsupervised LC SNN using a current-based LIF model, a simplified typical STDP, and the alternative version of competitive learning (Fig. 2(b)). This SNN is the first one to incorporate LC architecture. For a fair comparison, we strictly followed [8] to implement this method in the experiments, except that the size of input layer was changed from 20×20 to 28×28.

The two baseline methods above are chosen because the three of us (i.e., Diehl-FC, Saunders-LC, and ours) all focused on the design of network architecture. Diehl-FC and Saunders-LC represent the widely used FC architecture and more advanced LC architecture respectively. Other unsupervised SNNs (e.g., [9-10]) aren't our opponents because they didn't innovate architecture, and their contributions (e.g., learning theory, advanced STDP rules, etc.) can be assembled with our network architecture as well.

We also analyzed the effectiveness of our three contributions by ablation studies. We evaluated the performance degradation when each contribution was removed or replaced by an existing method as follows:

- Ours-FC: Our Inception-like architecture is replaced by the FC architecture.
- Ours-LC: Our Inception-like architecture is replaced by the LC architecture.
- Ours-noVFA: The VFA decoding layer is replaced by the standard vote-based spike decoding scheme.
- Ours-noAR: The Adaptive Repolarization (AR) mechanism is removed, so the LIF model with a fixed resetting voltage $v_{reset}$ are used to model output neurons.

More details about the baseline methods are shown in Table 1.

### 5.3 Training Procedure

We used a training procedure similar to the one used in [7]: In each iteration, we presented an image in the training set to the network for 350ms, and then there was a 150ms phase without any input to allow all variables of all neurons to decay to their default values. We trained our SNNs with a single pass through the training set, which leaded to 60,000/124,800 training iterations in total (one image per iteration) for MNIST/EMNIST. After the training was done, we set the learning rate to zero and fixed all synaptic weights. The synaptic weights of the VFA decoding layer were calculated based on the output neurons' spiking responses to the last 10,000 training images. Besides, we adopted the weight normalization scheme used in [8]: After each iteration, the sum of synaptic weights incident to an output neuron was normalized to be equal to a normalization constant $c_{norm}$. More implementation details including our code are presented in Appendix A1.

Table 1. Details of Baseline/Our Methods

| SNN | Architecture | Spike Decoding | Adaptive Repolarization | Competitive Learning | Neuron Model | STDP Model |
|---|---|---|---|---|---|---|
| Diehl-FC [7] | FC | Standard vote-based decoding [7] | no | Preliminary version (Fig. 2(a)) | conductance-based LIF [7] | Triplet STDP [35] |
| Saunders-LC [8] | LC | 2-gram [8] | no | Alternative version (Fig. 2(b)) | current-based LIF [8] | Simplified typical STDP [8] |
| Ours | Ours (Fig. 3) | VFA decoding layer | yes | Alternative version (Fig. 2(b)) | conductance-based LIF (1)-(6) | Simplified triplet STDP (7) |
| Ours-FC | FC | VFA decoding layer | yes | Alternative version (Fig. 2(b)) | conductance-based LIF (1)-(6) | Simplified triplet STDP (7) |
| Ours-LC | LC | VFA decoding layer | yes | Alternative version (Fig. 2(b)) | conductance-based LIF (1)-(6) | Simplified triplet STDP (7) |
| Ours-noVFA | Ours (Fig. 3) | Standard vote-based decoding [7] | yes | Alternative version (Fig. 2(b)) | conductance-based LIF (1)-(6) | Simplified triplet STDP (7) |
| Ours-noAR | Ours (Fig. 3) | VFA decoding layer | no | Alternative version (Fig. 2(b)) | conductance-based LIF (1)-(6) | Simplified triplet STDP (7) |

# 6 Evaluation

## 6.1 Learning Capability

In Table 2, we report the classification accuracy on the testing set of MNIST/EMNIST and the number of the neurons/synapses used in SNNs ($n_{neuron}/n_{synapse}$). Here we use $(k, s) \times F$ to denote the kernel size ($k$), stride ($s$), and feature map number ($F$) of a FC/LC architecture. As is mentioned in Section 4.1, each output neuron in the FC architecture can be regarded as a feature map, so the $(k, s)$ of a FC architecture is $(28,1)$. As shown in Table2, our SNNs achieved much improved testing results. Even more than 10% improvement on the EMINST was achieved by our SNNs. Besides, the SNNs with a larger scale normally exhibit better performance, while, compared to Diehl-FC and Saunders-LC, our SNNs achieved higher testing results using fewer neurons and synapses (Fig. 5). Note that we didn't test our SNN with $Size_{SA}$>400 because the SNNs with $Size_{SA} \leqslant 400$ have exhibited superior performance, but we anticipate that a larger $Size_{SA}$ will lead to a better result if computing resources are sufficient.

Table 3 shows the testing results of the ablation study where each our contribution was removed respectively. The experimental results show that: 1) Ours-FC and Ours-LC degraded in testing result, which further demonstrates that our proposed architecture outperforms the FC/LC architecture on learning capability; 2) Our-noAR exhibited lightly improved testing results, which suggests that the adaptive repolarization mechanism only has a negligible impact on learning capability but can obviously improve the SNNs' learning speed (shown in Section 6.2) and, 3) The testing results of Our-noVFA declined dramatically to even lower than Diehl-FC. This is because the standard vote-based decoding scheme only works in the FC architecture, while, in other architectures (e.g., LC, ours), it will cause a large information loss in spike decoding. Our architecture has superior learning capability, but this capability can't be reflected without the VFA decoding layer. This part will be further discussed in Section 7.2.

Table 2. Comparison of Learning Capability

| SNN | Topology | $n_{neuron}$ | $n_{synapse}$ | MNIST | EMNIST |
|---|---|---|---|---|---|
| Diehl-FC | $(28,1) \times 400$ | 800 | 473K | 87.88% | 66.41% |
| | $(28,1) \times 800$ | 1600 | 1267K | 90.22% | 67.45% |
| | $(28,1) \times 1600$ | 3200 | 3814K | 91.96% | 68.23% |
| | $(28,1) \times 6400$ | 12800 | 45977K | 94.97% | 47.41%* |
| Saunders-LC | $(16,6) \times 100$ | 900 | 320K | 91.36% | 62.37% |
| | $(16,6) \times 400$ | 3600 | 2361K | 93.97% | 66.78% |
| | $(16,6) \times 800$ | 7200 | 7603K | 94.83% | 68.83% |
| | $(16,6) \times 1000$ | 9000 | 11304K | 95.02% | 69.68% |
| Ours | $Size_{SA} = 100$ | 2100 | 1214K | 93.16% | 73.93% |
| | $Size_{SA} = 200$ | 4200 | 2849K | 94.19% | 76.45% |
| | $Size_{SA} = 300$ | 6300 | 4904K | 94.95% | 78.76% |
| | $Size_{SA} = 400$ | 8400 | 7379K | 95.64% | 80.11% |

* The training is divergent, and the highest result before divergence is listed.

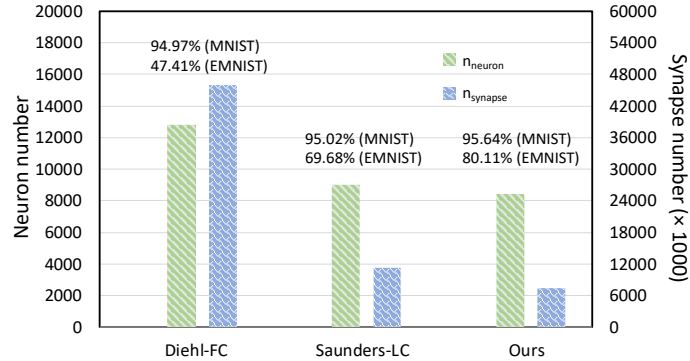

Fig. 5. A bar plot illustrating the number of the neurons/synapses ($n_{neuron}/n_{synapse}$) used in Diehl-FC, Saunders-LC, and our SNN. The testing accuracy is written above the bars.

Table 3. Testing Results of Ablation Study

| SNN | Topology | MNIST | EMNIST |
|---|---|---|---|
| Ours | $Size_{SA} = 400$ | 95.64% | 80.11% |
| Ours-FC | $(28,1) \times 6400$ | 95.06% | 56.16%* |
| Ours-LC | $(16,6) \times 1000$ | 95.26% | 77.85% |
| Our-noAR | $Size_{SA} = 400$ | 95.67% | 80.21% |
| Our-noVFA | $Size_{SA} = 400$ | 94.88% | 69.27% |

\* The training is divergent, and the highest result before divergence is listed.

Table 4. Comparison of Testing Results on the MNIST Dataset among the Existing Unsupervised SNN Algorithms

| Paper | Description | Result |
|---|---|---|
| Diehl et al. 2015 [7] | FC SNN (Diehl-FC) | 95.00% |
| Saunders et al. 2019 [8] | LC SNN (Saunders-LC) | 95.07% |
| Panda et al. 2017 [9] | FC SNN with Adaptive Synaptic Plasticity (ASP) | 96.80% |
| She et al. 2019 [10] | FC SNN with stochastic STDP | 96.10% |
| Rathi et al. 2018 [11] | Sparsely connected SNN with STDP-based connection pruning | 90.10% |
| Lammie et al. 2018 [12] | FPGA neuromorphic system based on FC SNN | 94.00% |
| Allred et al. 2016 [13] | FC SNN using forced firing of dormant or idle neurons | 85.90% |
| Querlioz et al. 2011 [14] | Memristor-Based SNN | 93.50% |
| Ours | High-parallelism Inception-like SNN | **95.64%** |

In Table 4, we compared our method with the various existing unsupervised SNN algorithms on the MNIST dataset. The results of the comparison algorithms were derived from the corresponding references. Note that semi-supervised SNN algorithms (e.g., [27-30, 41-42]) are excluded because they require extra supervised classifiers. Our method reached a superior performance than most of the comparison algorithms. Panda et al. [9] and She et al. [10] achieved higher results using more advanced learning rules (ASP, stochastic STDP), but they are still limited by the slow-learning FC architecture. Our contributions are mainly on architecture design and decoding mechanism, so other SNN components (e.g., neuron model, learning rules, etc.) used in the experiments are kept similar with Diehl-FC and Saunders-LC for a fair comparison. Nevertheless, our contributions are flexible and can work independently, which suggests that they can be in conjunction with other more advanced SNN components to achieve higher performance.

## 6.2 Learning Efficiency

In Fig. 6, we report the MNIST testing results of the SNNs trained with varying number of training iterations. In Fig. 6(a), we compared the SNNs with similar learning capability: Diehl-FC with $(28,1) \times 6400$, Saunders-LC with $(16,6) \times 800$, and our SNN with $Size_{SA} = 300$. These three SNNs all exhibited close learning capabilities of about 94.9% accuracy when fully trained. It's shown in Fig. 6(a) that our SNN exhibited much superior learning efficiency. The testing results of our SNN can achieve nearly 80%, 90% accuracy with only 500, 2500 training iterations, while Saunders-LC just achieved nearly 30%, 75% accuracy with 500, 2500 training iterations, and worst of all, Diehl-FC only reached nearly 42% accuracy with even 10,000 training iterations. Note that Saunders-LC and our SNN can be fully trained within one pass through the training set, but Diehl-FC needs to be trained with 15 passes through the training set, resulting in 900,000 training iterations in total.

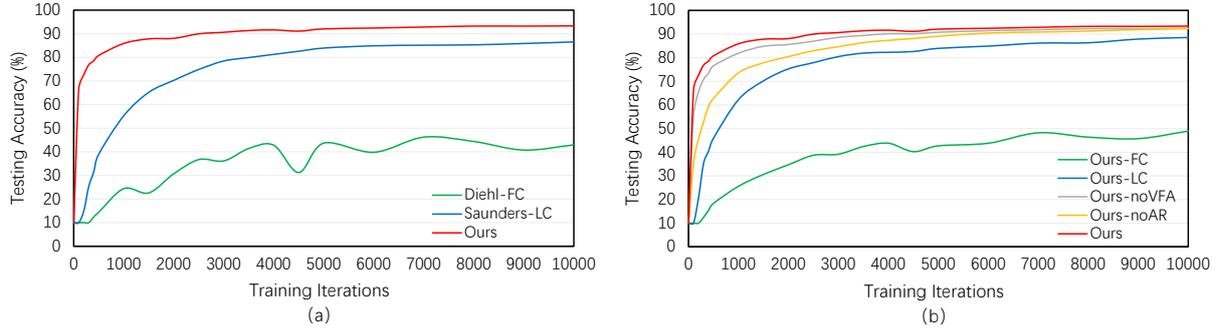

Fig. 6. Testing results of the SNNs trained with varying number of training iterations. (a) Comparison among Diehl-FC, Saunders-LC, and ours. (b) Ablation study.

Fig. 6(b) shows the results of the ablation study where Ours-FC with $(28,1) \times 6400$, Ours-LC with $(16,6) \times 800$, and Ours/Ours-noVFA/Ours-noAR with $Size_{SA} = 300$ were tested for learning efficiency. The experimental results show that: 1) Ours-FC and Ours-LC exhibited much slower learning speed than our original SNN, which demonstrates that our proposed architecture outperforms the FC/LC architecture on learning efficiency; 2) The learning efficiency of Our-noAR also degraded, meaning that the adaptive repolarization mechanism is capable of accelerating our SNN's learning and, 3) The testing curve of Our-noVFA is below the one of our original SNN, but they have a similar growing pace, which suggests that the VFA decoding layer improves our SNN's inference accuracy but doesn't have any obvious impact on our SNN's learning efficiency.

### 6.3 Robustness

Following Saunders et al. [8], by 'Robustness' we mean SNNs' resistance against hardware damage and destruction. To evaluate it, we followed the experiments used in [8]. This experiment simulates a situation when SNN-based hardware gets physical damage or even destructed. Concretely, we randomly deleted the output neurons or learnable synapses of trained SNNs with probability $\rho_{delete}$, then report their testing results on the MNIST dataset in Fig. 7.

Fig. 7(a)/(b) shows the experimental results of Diehl-FC with $(28,1) \times 6400$, Saunders-LC with $(16,6) \times 800$, and our SNN with $Size_{SA} = 300$. The three SNNs without any neuron/synapse deletion have similar learning capability, all reaching a result of about 94.9%, while deleting neuron/synapse leads to performance degradation. Fig. 7(a) shows the testing results after deleting neurons with probability $\rho_{delete}$, and our SNN exhibited the highest robustness: Our SNN achieved nearly 90%, 80% accuracy even if 90%, 95% of neurons were deleted, while Diehl-

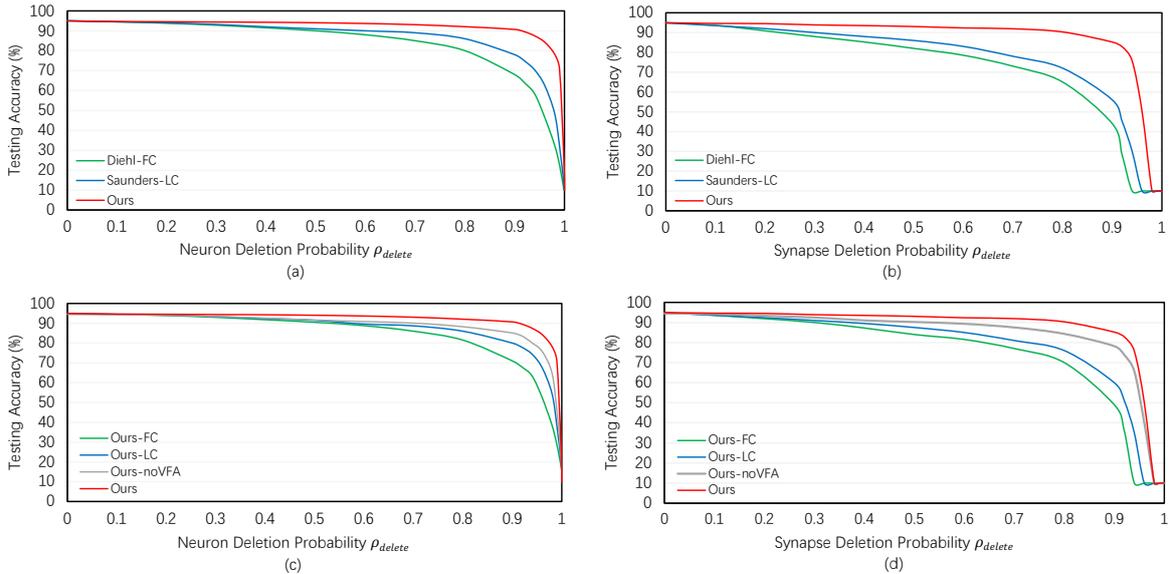

Fig. 7. Testing results of the SNNs whose neurons/synapses are randomly deleted with varying probability $\rho_{delete}$. (a) Neuron deletion. (b)Synapse deletion. (c)/(d) Ablation studies for neuron/synapse deletion.

FC and Saunders-LC achieved nearly 80% and 85% accuracy, respectively, with only 80% of neurons deleted. Similarly, Fig. 7(b) shows the testing results after deleting synapses with probability $\rho_{delete}$, and our SNN still exhibited the highest robustness: Our SNN achieved nearly 90%, 80% results even if 80%, 90% of synapses were deleted, while Diehl-FC and Saunders-LC just achieved nearly 70% and 65% results with only 80% of synapses deleted. This experiment demonstrates that our SNN have higher robustness against hardware damage and can work well even when most of its learnable synapses or computing neurons have broken down.

Also, we tested Ours-FC with $(28,1) \times 6400$, Ours-LC with $(16,6) \times 800$, and Ours/Ours-noVFA with $Size_{SA} = 300$ in Fig. 7(c)/(d). The experimental results show that: 1) Ours-FC and Ours-LC exhibited much degraded robustness compared to our original SNN, which demonstrates that our proposed architecture outperforms the FC/LC architecture on robustness and, 2) Surprisingly, Ours-noVFA also exhibited degraded robustness, especially in Fig. 7(c) (neuron deletion), which suggests that the VFA decoding layer helps improving SNNs' robustness. This is an unintended benefit because the VFA decoding layer was originally designed to reduce the information loss in spike decoding and to improve inference accuracy. Detailed discussion is shown in Section 7.2. Note that the curve of Ours-noAR is not drawn in Fig. 7(c)/(d) because the curve of Ours-noAR is almost the same as the one of our original SNN. This suggests that our adaptive repolarization mechanism only has a negligible impact on SNNs' robustness.

## 7 Discussion

In Section 7.1, we analyze the parallelism of FC/LC/our architecture. Our architecture consists of 21 sub-networks that can learn and compute in parallel, which explains our SNN's improved learning efficiency and robustness. In Section 7.2, the information loss in spike decoding is analyzed. Our VFA decoding layer matches the learning nature of our architecture, thus resulting in less information loss. In Section 7.3, we discuss the relationship between spiking intensity and learning efficiency. The spiking intensity is tested to demonstrate that our adaptive repolarization mechanism can enhance the spiking activities of winner neurons, and thereby accelerates learning.

### 7.1 Architecture Parallelism

A SNN's learning efficiency and robustness against hardware damage is highly relevant to its architecture parallelism. We assumed that, if a SNN consists of multiple independent sub-networks and these sub-networks can compute in parallel, this SNN can exhibit superior learning efficiency and robustness because all sub-networks can learn simultaneously and work independently when other sub-networks get damaged or even destructed.

In competitive learning, as we mentioned in Section 3.5, a competition area is defined as a set of output neurons sharing the same RF, and each output neuron only competes with other neurons in the same competition area. According to this principle, a competition area can be regarded as a sub-network, because there is no interaction between two competition areas and each competition area can compute independently. Fig. 8 shows the partitions of FC/LC/our SNN's competition areas (sub-networks). The widely used FC architecture has only a single competition area because all output neurons share the same global RF (Fig. 8(a)). A large number of output neurons are located in one competition area, leading to sub-optimal learning efficiency and robustness. The LC architecture eased this limitation by localizing the RF (Fig. 8(b)), which allows multiple competition areas (sub-networks). Therefore, the LC architecture can exhibit improved learning efficiency and robustness. We took a further step based on the LC architecture: 1) Inspired by the Inception module in ANN literature, we designed an Inception-like multi-pathway architecture to integrate multi-scale spatial information and improve the network's parallelism and, 2) We partitioned the original competition area into multiple sub-areas to further increase the number of competition areas. As shown in Fig. 8(c), our architecture has 21 competition areas (sub-networks) in total, which explains why our architecture can outperform the FC/LC architecture in terms of learning efficiency and robustness.

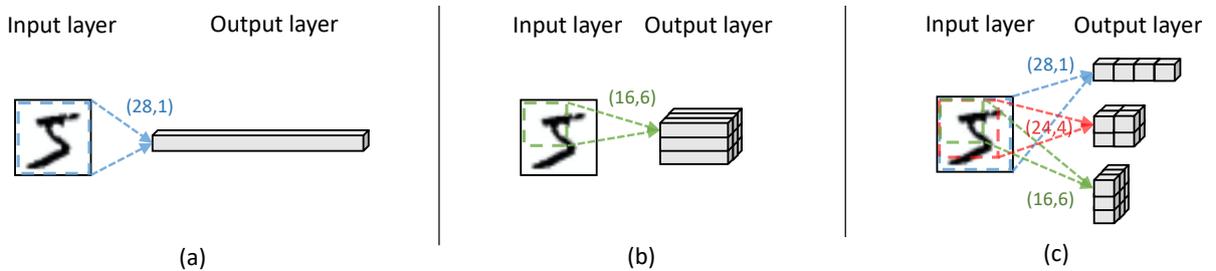

Fig. 8. An illustration of SNN's competition areas. (a) FC architecture contains only a single competition area. (b) LC architecture contains 9 competition areas. (c) Our architecture contains 21 competition areas. Note that a cube denotes a competition area here.

## 7.2 Information Loss in Spike Decoding

In the experiments, we found that the standard vote-based decoding scheme (denote by VFO here) worked well in the FC architecture but performed poorly in the LC/our architecture (Fig. 10). We attribute this because one underlying assumption of the VFO is violated: The VFO assigns each output neuron to a certain class, and builds a Vote-for-One (VFO) relationship between each output neuron and a certain class. Under this strategy, each output neuron can only vote for one certain class. This approach is based on an assumption that each output neuron is only highly active to one certain class. This assumption is feasible in the FC architecture because each output neuron can receive global information (Fig. 9(a)). This assumption, however, is violated in the LC architecture because the output neurons in this architecture can only receive local information (Fig. 9(b)). Since the images belonging to different classes might have similar local features, theoretically an output neuron in the LC architecture can be highly active to multiple classes. For example, as shown in Fig. 9(b), the local features of digits '2' and '3' in the red square are quite similar, so the output neurons corresponding to these RFs might be highly active to both '2' and '3' because they cannot classify them based on the local features. Violating this underlying assumption leads to a large information loss in spike decoding. Similarly, the VFO is not applicable to our architecture due to the same reason. In contrast to the VFO, the proposed VFA decoding layer (denoted by VFA here) builds a Vote-for-All (VFA) relationship between each output neuron and all possible classes, and thus allows each output neuron to vote for all classes, which greatly reduces the information loss in spike decoding.

To validate our above hypothesis, in Fig. 10, we compared the performance improvements when replacing VFO with VFA in three network architectures (FC with $(28,1) \times 6400$, LC with $(16,6) \times 1000$, and Ours with $Size_{SA} = 400$) on the MNIST. For a fair comparison, these three SNNs were implemented using the same LIF model, same STDP rules, and the alternative version (Fig. 2(b)) of competitive learning. The adaptive repolarization was not used here. The experimental results show that the LC/our architecture gained more performance improvement than the FC architecture. This suggests that, without VFA, LC/our architecture cannot fully exhibit its superior learning capability. We suppose that, in previous studies, scholars might have tried to use Inception-like architecture as well, but they failed to get improvement without VFA.

Moreover, in Fig.7(c)/(d), we surprisingly found that the VFA decoding layer helped to improve our SNN's robustness. The possible explanation is that, under the VFO approach, each class can only get votes from the output neurons assigned to this class, while, under the VFA approach, each class can get votes from all output neurons. In the latter case, even many broken neurons won't have a fatal impact on the final inference result because the remaining output neurons can still work together to infer a reliable result.

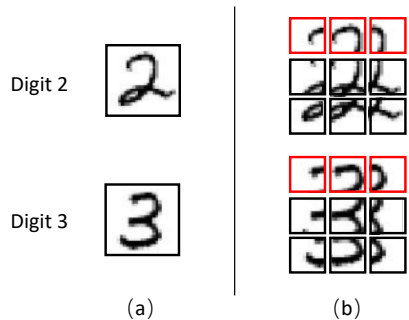

Fig. 9. An example of output neurons' RFs in the (a) FC and (b) LC architectures when input images are digit '2' or '3'. The output neurons in the (a) can see the whole input images, while the ones in the (b) only can see a part of the input images.

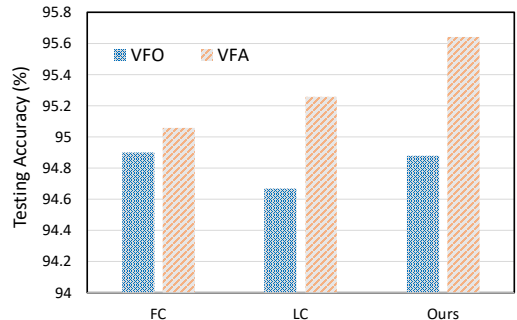

Fig. 10. A bar plot illustrating the improvements in testing results when VFO is replaced with VFA in the FC/LC/Our architecture.

## 7.3 Spiking Intensity and Learning Efficiency

Theoretically, the learning efficiency of a STDP-based SNN is highly relevant to its spiking intensity, because, according to the STDP rules, the update of synaptic weights (i.e., learning) occurs when neurons fire spikes. To validate this hypothesis, we tested the average spiking intensity of Diehl-FC with $(28,1) \times 6400$, Saunders-LC with $(16,6) \times 1000$, and our SNN with $Size_{SA} = 400$, and report the results in Table 5. The experimental results show that the Saunders-LC and our SNN have higher spiking intensity than Diehl-FC, and ours has the highest spiking intensity. These results conform to the hypothesis that the SNN with higher spiking intensity exhibits faster learning speed. This also allows us to explain why our SNN can learn faster from another point of view.

The effect of our adaptive repolarization mechanism on learning capability and learning efficiency is shown in Fig. 11, where we tested the average spiking intensity of our SNN with varying parameter $\alpha$ of adaptive repolarization on the MNIST dataset. Note that, in this experiment, the $Size_{SA}$ was set to be 400, and the $\alpha$ was fixed once it's set at the beginning of the training. Here we use the testing result of the SNN trained with 500 iterations to represent the learning efficiency, and use the testing result of the fully trained SNN to represent the learning capability. As shown in Fig. 11, there is a trade-off between learning capability and learning efficiency: A larger $\alpha$ leaded to higher spiking intensity and higher learning efficiency but lower learning capability. This experiment further demonstrates the relationship between learning efficiency and spiking intensity and also explains why we used a declined $\alpha$ in our training procedure. Note that, as we mentioned in Section 4.3, the adaptive repolarization mechanism can enhance the spiking activities of the winner neurons but hinder the ones of other neurons. However, since SNNs' spiking intensity is dominated by winner neurons, we only observed an increase of spiking intensity when the adaptive repolarization mechanism is used.

Table 5. Comparison of Average Spiking Intensity

| SNN | Spiking Intensity (spikes/iteration) |
|---|---|
| Diehl-FC | 7.84 |
| Saunders -LC | 58.26 |
| Ours | 154.63 |

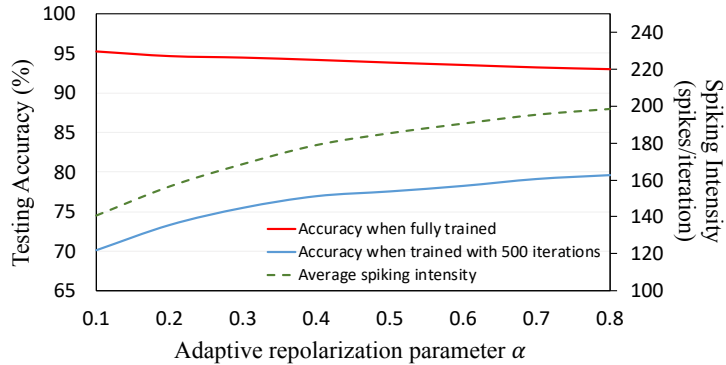

Fig. 11. Testing results and average spiking intensity of our SNN with varying $\alpha$ of adaptive repolarization.

## 8 Conclusion and Future Work

We proposed a fast-learning and high-robustness unsupervised SNN using a high-parallelism Inception-like network architecture. In our experiments, our architecture outperformed the widely used FC architecture and more advanced LC architecture, and our SNN achieved improvements in all learning capability, learning efficiency, and robustness against hardware damage. Moreover, the proposed VFA decoding layer was recognized in reducing the information loss in spike decoding.

Our method can be further improved. To emphasize the effectiveness of our contributions, other SNN components used in our experiments are relatively simple to the optimized state-of-the-art counterparts. We suggest that more advanced SNN components (e.g., ASP [9], stochastic STDP [10], etc.) can be used to achieve better performance. Besides, although we used our Inception-like architecture on unsupervised SNNs, there is a possibility of it being added to supervised SNNs because our architecture can integrate multi-scale information (features), which is helpful for supervised feature learning as well.

# Appendix

## A.1 Implementation Details

Our experiments ran on an Ubuntu system with Python 2.7. Our code is based on an open-source simulator, Brian [37], and is available at *https://github.com/MungoMeng/Spiking-Inception*.

All hyperparameters used in our experiments are empirical values (in Table 6). The $\alpha$ of the adaptive repolarization mechanism is 0.6 initially, and then halves every 5000 iterations. Finally, the $\alpha$ is set to 0 after the 20,000[th] training iteration. We decided the hyperparameters through cross-validation, in which we randomly picked up 10,000/20,800 images from the training set of MNIST/EMNIST as the validation set.

Table 6. Hyperparameter Settings in the Experiments

| Hyperparameter | Description | Value |
|---|---|---|
| $\eta_{post}$ | Postsynaptic learning rate | 0.01 |
| $\eta_{pre}$ | Presynaptic learning rate | 0.0001 |
| $\tau_{pre}$ | Time constant of $x_{pre}$ | 20ms |
| $\tau_{post1}/\tau_{post2}$ | Time constant of $x_{post1}/x_{post2}$ | 20ms/40ms |
| $v_{thres}$ | Threshold voltage | -52mv |
| $v_{rest}$ | Resting voltage | -65mv |
| $v_{reset}$ | Reseting voltage | -65mv |
| $v_{exc}/v_{inh}$ | Equilibrium voltage | 0mv/-100mv |
| $\theta_{plus}$ | Increment for adaptive threshold | 0.05mv |
| $T_{ref}$ | Time length of refractory period | 5ms |
| $\tau_{v}$ | Time constant of $v$ | 100ms |
| $\tau_{\theta}$ | Time constant of adaptive $\theta$ | $10^{7}$ms |
| $\tau_{g_e}/\tau_{g_i}$ | Time constant of $g_e/g_e$ | 1ms/2ms |
| $c_{norm}$ | Weight normalization constant | 78.4 |
| $\mu$ | VFA decoding layer constant | 0.1 |

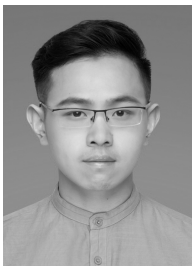

**Mingyuan Meng** received the B.E. degree in electronic information science and technology from Tsinghua University, Beijing, China, in 2018. He is currently pursuing a Ph.D. degree with the School of Computer Science, the University of Sydney, Sydney, Australia.

He was a research assistant at the School of Electronics and Information Technology, Sun Yat-Sen University, Guangzhou, China, in 2019. His current research interests include machine learning, artificial/spiking neural networks, and medical image analysis.

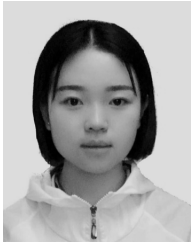

**Xingyu Yang** is currently pursuing the graduate degree with the School of Electronic Information and Engineering, Sun Yat-Sen University of China, Guangzhou, China, from 2019. Her current research interests include spiking neural networks, spiking neuron models, and neuromorphic hardware.

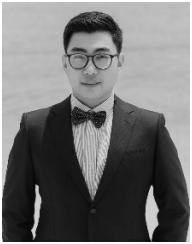

**Lei Bi** received his master of information technology, master of philosophy (research) and PhD from the University of Sydney, in 2011, 2013 and 2018, respectively.
Currently, he is a research fellow with the Australia Research Council Training Centre in Innovative BioEngineering, the University of Sydney. His research interests include in using deep learning technologies for computer-aided diagnosis and medical image analysis.

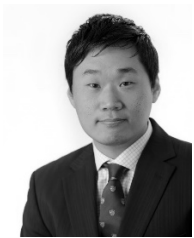

**Jinman Kim** received the B.S. (honours) degree in computer science and PhD degree from the University of Sydney, Australia, in 2001 and 2006, respectively.
Since his PhD, he has been a Research Associate at the leading teaching hospital, the Royal Prince Alfred. In 2008 until 2012, he was an ARC postdoc research fellow, one year leave (2009-2010) to join MIRALab research group, Geneva, Switzerland, as a Marie Curie senior research fellow. Since 2013, he has been with the School of Computer Science, The University of Sydney, where he was a Senior Lecturer, and became an Associate Professor in 2016. His research interests include medical image analysis and visualization, computer aided diagnosis, and Telehealth technologies.

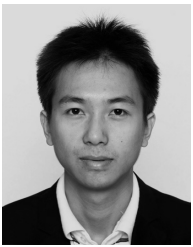

**Shanlin Xiao** received the B.S. degree in communications engineering and the M.S. degree in communications and information systems from the University of Electronic Science and Technology of China (UESTC), Chengdu, China, in 2009 and 2012, respectively. He received his Ph.D. degree in Communications and Computer Engineering from the Tokyo Institute of Technology, Tokyo, Japan, in 2017.
He is currently an associate research professor at the School of Electronics and Information Technology in Sun Yat-Sen University, Guangzhou, China. His research interests include domain-specific architecture for artificial intelligence and neuromorphic computing.

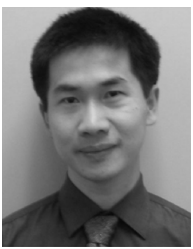

**Zhiyi Yu** received the B.S. and M.S. degrees in EE from Fudan University, China, in 2000 and 2003, respectively, and the Ph.D. degree in ECE from the University of California at Davis, CA, USA, in 2007.
He was with IntellaSys Corporation, CA, USA, from 2007 to 2008. From 2009 to 2014, he was an associate professor in the Department of Microelectronics, Fudan University, China. Currently, he is a professor at the school of electronics and information technology, Sun Yat-sen University, China. His research interests include digital VLSI design and computer architecture. Dr. Yu serves as TPC member on many conference committees, such as the ASSCC, VLSI-SOC, ISLPED, APSIPA, SASIM.